%% file: main.tex
\documentclass{article}

\usepackage{acro}
\acsetup{
  list/heading   = section,
  barriers/use   = true
}
\usepackage{booktabs}     
\usepackage{multirow}
\usepackage{makecell} 
\usepackage{etoolbox}
\AfterEndEnvironment{abstract}{\acbarrier}
\usepackage{multirow}
\usepackage[preprint]{custom}        
\usepackage[numbers,sort&compress,square]{natbib}
\usepackage[utf8]{inputenc}
\usepackage[T1]{fontenc}
\usepackage{newtxtext,newtxmath}     
\usepackage{graphicx}
\usepackage{xcolor}
\usepackage{hyperref}
\usepackage{url}
\usepackage{booktabs}
\usepackage{amsfonts}
\usepackage{nicefrac}
\usepackage[protrusion=true,expansion=false]{microtype}
\usepackage{subcaption}
\usepackage{acro}  
\usepackage{listings}
\usepackage{booktabs, siunitx, makecell}
\usepackage{booktabs} 
\usepackage{multirow}

\usepackage{booktabs, makecell, siunitx}
\input{acronyms.tex} 

\usepackage{etoolbox}
\AfterEndEnvironment{abstract}{\acbarrier}

\title{Arrow-Guided VLM: Enhancing Flowchart Understanding via Arrow Direction Encoding}

\author{%
  Takamitsu Omasa \quad
  Ryo Koshihara \quad
  Masumi Morishige\\[2pt]
  Galirage Inc.\\[2pt]
  \texttt{info@galirage.com}
}

\date{\today}

\begin{document}
\maketitle

\input{sections/abstract.tex}

\acbarrier

\input{sections/introduction.tex}
\input{sections/relatedworks.tex}
\input{sections/methodology.tex}

\input{sections/results.tex}

\input{sections/discussion}
\input{sections/conclusion.tex}

\bibliographystyle{unsrtnat}
\bibliography{references}

\appendix
\input{sections/appendix.tex}

\end{document}

%% file: acronyms.tex
\DeclareAcronym{vlm}{
  short = VLM,
  long  = Vision Language Model,
  short-plural = s, 
  long-plural = s   
}

\DeclareAcronym{ocr}{
  short = OCR,
  long  = Optical Character Recognition
}

\DeclareAcronym{bpmn}{
  short = BPMN,
  long  = Business Process Model and Notation
}

\DeclareAcronym{uml}{
  short = UML,
  long  = Unified Modeling Language
}

\DeclareAcronym{llm}{
  short = LLM,
  long  = Large Language Model,
  short-plural = s,
  long-plural = s
}

\DeclareAcronym{mmlu}{
  short = MMLU,
  long  = Massive Multitask Language Understanding
}

\DeclareAcronym{mmmu}{
  short = MMMU,
  long  = Massive Multi-discipline Multimodal Understanding
}

\DeclareAcronym{docvqa}{
  short = DocVQA,
  long  = Document Visual Question Answering
}

\DeclareAcronym{gpt4v}{
  short = GPT-4V,
  long  = GPT-4 with Vision
}

\DeclareAcronym{sam}{
  short = SAM,
  long  = Segment Anything Model
}

\DeclareAcronym{vqa}{
  short = VQA,
  long  = Visual Question Answering
}

\DeclareAcronym{ai2d}{
  short = AI2D,
  long  = AI2 Diagrams
}

\DeclareAcronym{scrm}{
  short = SCRM,
  long  = Structured Chart Representation Matching
}

\DeclareAcronym{em}{
  short = EM,
  long  = Exact Match
}

\DeclareAcronym{iou}{
  short = IoU,
  long  = Intersection-over-Union
}

\DeclareAcronym{ap}{
  short = AP,
  long  = Average Precision
}

\DeclareAcronym{ar}{
  short = AR,
  long  = Average Recall
}

\DeclareAcronym{map}{
  short = mAP,
  long  = mean Average Precision
}

%% file: sections/abstract.tex
\begin{abstract}
Flowcharts are indispensable tools in software design and business-process analysis, yet current \acp{vlm} frequently misinterpret the directional arrows and graph topology that set these diagrams apart from natural images.
This paper introduces a seven-stage pipeline, grouped into three broader processes—(1) arrow-aware detection of nodes and arrow endpoints; (2) \ac{ocr} to extract node text; and (3) construction of a structured prompt that guides the \acp{vlm}.
Tested on a 90-question benchmark distilled from 30 annotated flowcharts, our method raises overall accuracy from 80\% to 89\% (+9 pp), a sizeable and statistically significant gain achieved without task-specific fine-tuning of the \acp{vlm}.
The benefit is most pronounced for next-step queries (25/30 → 30/30; 100\%, +17 pp); branch-result questions improve more modestly, and before-step queries remain difficult.
A parallel evaluation with an LLM-as-a-Judge protocol shows the same trends, reinforcing the advantage of explicit arrow encoding.
Limitations include dependence on detector and \ac{ocr} precision, the small evaluation set, and residual errors at nodes with multiple incoming edges.
Future work will enlarge the benchmark with synthetic and handwritten flowcharts and assess the approach on \ac{bpmn} and \ac{uml}.
\end{abstract}

%% file: sections/introduction.tex
\section{Introduction}
Flowcharts distill complex control flow, decision logic, and data transformations into a handful of boxes and arrows. In software engineering and business-process management, these diagrams are more than didactic artifacts, as such diagrams enable automatic code generation and serve as effective pedagogical tools \cite{Hooshyar_2015, Shukla2025}.

Within just three years, \acp{llm} have advanced at an unprecedented pace: accuracy on the 57-subject \ac{mmlu} suite climbed from 43.9\% with GPT-3 (2021) \citep{Hendrycks2020} to nearly 89\% with GPT-4o \citep{OpenAI2024}.
\acp{vlm} likewise achieve benchmark-leading results on diverse multimodal benchmarks; for instance, GPT-4o excels on \ac{mmmu}, MathVista, and \ac{docvqa} \citep{Hello-2025-04-30,Yue2023, Lu2023, Mathew2020}. However, its high accuracy deteriorates markedly once explicit graph-topology reasoning is required. On the simulated subset of the \textsc{FlowLearn} benchmark, converting flowcharts to Mermaid code still proves challenging: on the link-level F$_1$ metric, Claude-3 Opus scores 0.30 and \ac{gpt4v} only 0.22 (100-sample subset; \citealp{Pan_2024}, Table 7), underscoring how current \acp{vlm} struggle to recover edge relationships.

\begin{figure*}[t] 
  \centering
  \includegraphics[width=0.9\linewidth]{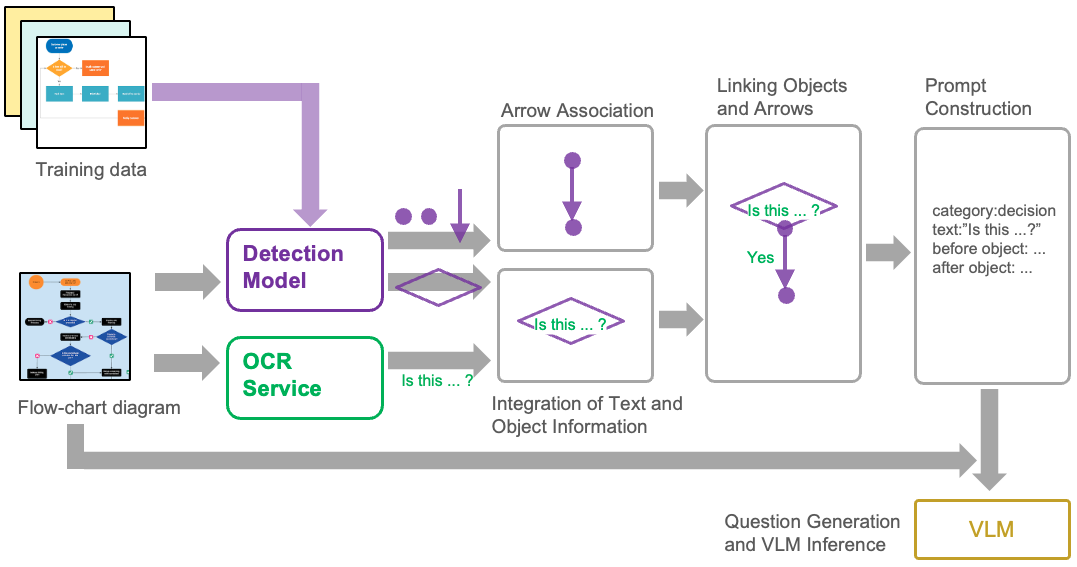}
  \caption{Overview of the seven-stage pipeline: OCR, object detection, text–object fusion, arrow anchoring, node–arrow linking, graph-structured prompt generation, and VLM-based reasoning.}
  \label{fig:pipeline}
\end{figure*}

Previous approaches can be categorized into two main types.
First, some studies couple off-the-shelf object detectors such as YOLO \citep{Redmon2015} with \ac{ocr}; the resulting bounding boxes and tokens are concatenated into a prompt for a \ac{vlm}, yielding only modest gains over detector-free baselines.  
Second, other work relies on zero-shot segmentation models, most prominently the \ac{sam} \citep{Kirillov2023}. GenFlowchart \citep{Arbaz_2024}, for instance, converts \ac{sam} masks into bounding boxes, adds \ac{ocr}, and queries  GPT-3.5 Turbo, yet still suffers from arrow-ordering ambiguities and localization noise.  
A complementary strand improves the detector itself—arrow-aware models like Arrow R-CNN halve localization errors on handwritten diagrams \citep{Sch_fer_2021}—but these specialised detectors have not been fused with \acp{llm}. Their outputs feed rule-based pipelines, so branch ordering and multi-step reasoning remain unresolved.
To close this gap, we propose the first detector–\ac{vlm} fusion pipeline for flowcharts (Fig.~\ref{fig:pipeline}). First, a fine-tuned, arrow-aware detector localises nodes and arrowheads. The \ac{ocr} stage then extracts textual labels. Finally, the {\emph{text}, \emph{bbox}} pairs are serialised into a coordinate-rich prompt that, together with the image, is fed to GPT-4o. Unlike prior work that lists raw labels, each token is annotated with its normalised center-of-mass, allowing the \ac{vlm} to infer edge orientations through the geometric priors internalised during pre-training.

Motivated by these gaps, this study investigates whether \emph{tightly coupling a flowchart-aware detector with a \ac{vlm} via a coordinate-rich prompt can close the reasoning gap on diagrammatic tasks.}  To investigate this question, an Arrow-aware detector is fused with GPT-4o and evaluated on a new 90-question suite spanning three query types and diverse diagram complexities, observing up to +9 pp overall and 100\% accuracy on next-step queries.

These gains rest on a relatively small test set—90 questions from 30 diagrams—and remain bounded by the detection model's localization accuracy.
These results are therefore viewed as a first step; scaling the benchmark, \emph{adapting the pipeline to large public corpora such as \textsc{FlowLearn}}, and exploring detector–\ac{vlm} co-training are left for future work.\cite{Pan_2024}

%% file: sections/relatedworks.tex
\section{Related Work}

\subsection{Limitations of End-to-End VLMs on Diagram Tasks}
VLMs such as GPT-4o achieve state-of-the-art scores on natural-image VQA and captioning benchmarks; however, their accuracy drops sharply when tasks demand explicit reasoning over graph topology or precise measurement rather than free-form visual cues. \cite{Hello-2025-04-30} 
\citep{Pan_2024} show that on the \textsc{FlowLearn} benchmark \textsc{GPT-4V} and \textsc{Claude-3} achieve only F\textsubscript{1}\,$=0.22$ and $0.30$, respectively, when translating simulated flowcharts into Mermaid code or answering edge-oriented questions; most errors stem from missed arrowheads, confusion between incoming and outgoing edges, and OCR noise that propagates through the reasoning process.

When \citet{Chen2024} re-evaluated the \ac{ai2d} corpus originally introduced by \citet{Kembhavi2016}, \ac{gpt4v} answered just 75.3\,\% of questions on the AI2D-Test split—well below human-level performance.  Chen \textit{et al.} attribute the gap to questions that can be solved without genuine visual reasoning and to potential data leakage, while follow-up error analyses in diagram-specific benchmarks (e.g., FlowLearn) highlight persistent failures to associate arrows, call-outs, and legend entries with their correct textual referents.

Data-visualisation benchmarks reinforce the trend.  
On the ChartInsights low-level ChartQA benchmark
\citep{Wu2024}, \ac{gpt4v} answers only
56.1\% of questions with a vanilla prompt
(rising to 66.4\% under a Yes/No prompt),
and simple corruptions—most notably median blur—degrade
accuracy by about 15 percentage points.

On the larger, real-world CharXiv corpus, \citet{Wang2024} shows that \textsc{GPT-4o} answers only 47.1\,\% of reasoning questions correctly.  Similarly, \citet{Xia2024} report that \ac{gpt4v} attains just 33\,\% accuracy on the ChartX question-answering task and no more than 27.2\,AP on the accompanying \ac{scrm} benchmark, which measures table reconstruction quality.

Across these datasets—flowcharts \cite{Pan_2024}, textbook illustrations \cite{Chen2024}, and statistical charts \cite{Wang2024, Xia2024} —four failure modes recur:  
(i) entity–label misalignment caused by invisible coordinates,  
(ii) cascading \ac{ocr} errors,  
(iii) ambiguity in arrow or series direction, and  
(iv) acute sensitivity to minor visual perturbations such as color-map changes or compression artefacts.  
A purely end-to-end multimodal transformer therefore lacks the \emph{geometry channel} required for reliable diagrammatic reasoning, motivating approaches that preserve spatial layout explicitly.

\subsection{Object-Detection–Driven Flowchart Interpretation}
Many studies mitigate \acp{vlm}' topological blind spots via a two-stage recipe: \emph{first localize the entities, then let the language model reason}.  
The \textsc{FlowLearn} baseline exemplifies this design: a detector–plus–\ac{ocr} front-end extracts node boxes and labels, which are concatenated into a prompt for \ac{gpt4v}; node-level detection is accurate, yet edge-level F\textsubscript{1} drops to~0.22 because the prompt conveys no spatial cues~\citep{Pan_2024}.  
GenFlowchart strengthens the vision stage by replacing the task-specific detector with the zero-shot \ac{sam} proposed by \citet{Kirillov2023}. \ac{sam}’s universal masks are collapsed to bounding boxes, optical character recognition is applied, and the resulting \{\emph{mask}, \emph{text}\} pairs are forwarded to GPT-3.5-Turbo, following the pipeline of \citet{Arbaz_2024}. Although this design boosts embedding-based textual-similarity scores, our replication shows that it still misorders branches whenever two nodes share the same axis—a structural error the original paper does not report.  
For hand-drawn sketches, \citet{Sch_fer_2021} introduces Arrow R-CNN, which augments Faster\,R-CNN with head–tail keypoint predictors and halves localization error on four datasets, but its output flows into a rule-based graph builder rather than a modern \acp{vlm}.  

What unites these pipelines is the disappearance of the \emph{geometry channel}: bounding-box centers, pairwise distances, and arrow orientations are either discarded or embedded latently, so the language model must hallucinate topology from an unordered token list.  Work on natural images confirms that explicit coordinates can help—Shikra encodes clicked points as textual tags~\citep{Chen2023}, ChatSpot leverages instruction tuning for precise region references~\citep{Zhao2023}, and RegionBLIP injects positional features as soft prompts~\citep{Zhou2023}—yet none of these systems target graph-based diagrams such as flowcharts.

A separate research line removes the interface altogether by predicting structure end-to-end. GRCNN outputs node categories and an adjacency matrix in a single forward pass before emitting code with a syntactic decoder~\citep{Cheng2020}.  FloCo-T5 is trained on 11,884 flow-chart images and surpasses a vanilla CodeT5 baseline with 67.4 BLEU, 75.7 CodeBLEU, and 20 \% \ac{em} \citep{Shukla2025}.
The authors also show a sharp drop to 21.4 BLEU on 40 hand-drawn diagrams, indicating limited robustness to noisy or off-distribution inputs.
Because FloCo-T5 directly decodes a fixed "FloCo" token stream into Python, it has not yet been evaluated for integrating external knowledge or chain-of-thought reasoning (our observation).

Previous work splits into two extremes: (i) detector-plus-\ac{llm} pipelines that drop coordinates before reasoning, and (ii) end-to-end models that predict the full graph in one shot but sacrifice linguistic flexibility. We introduce the first pipeline that retains every entity as a (text, x, y) tuple and feeds this sequence directly to \ac{vlm}, closing the gap between spatial fidelity and expressive reasoning.

%% file: sections/methodology.tex
\section{Methodology}

Our proposed inference pipeline comprises seven sequential stages: text extraction via \ac{ocr}, object detection, integration of text and objects, association of arrows with their start and end points, linking objects to arrows, prompt construction reflecting graph structure, and finally, question generation and \ac{vlm}-based reasoning.  
The overall architecture is illustrated in Figure~\ref{fig:pipeline}.

\subsection{Text Extraction via \ac{ocr}}

First, we apply the Azure AI Document Intelligence service to each input flowchart image to extract textual content and corresponding bounding box coordinates. Off-the-shelf \ac{ocr} tools were used without modification, leveraging their robust performance on printed and scanned text. The extracted texts and their spatial locations form the initial input to the downstream processes.

\subsection{Object Detection}

We then detect key flowchart elements—such as processes, decisions, and arrows—using a fine-tuned object-detection model. Specifically, we adopt the DAMO-YOLO model~\cite{Xu2022}, which is distributed under the Apache 2.0 license and delivers competitive accuracy comparable to state-of-the-art detection models.

We annotate nine object classes within the flowcharts:
\begin{enumerate}
    \item Text
    \item Arrow
    \item Terminator
    \item Data
    \item Process
    \item Decision
    \item Connection
    \item Arrow Start
    \item Arrow End
\end{enumerate}

For classes 1--7, standard bounding boxes encapsulate the relevant regions. For \textit{Arrow Start} and \textit{Arrow End}, we annotate small bounding boxes tightly around the visual start and end points of each arrow, respectively. It is noteworthy that arrows themselves can sometimes span very large bounding boxes, reflecting their visual prominence.

Although text was initially annotated as an object, in the final implementation, we instead relied exclusively on the coordinates obtained from the \ac{ocr} service for text information. Of the total 99 annotated diagrams, 30 were reserved for testing, while the remaining 69 were used for training and validation.

\subsection{Integration of Text and Object Information}

Next, we merge the \ac{ocr}-derived text information with the detected object information. We exclude arrows (\textit{Arrow}, \textit{Arrow Start}, and \textit{Arrow End}) from this integration step.

For each text bounding box, if it overlaps by more than 50\% with a detected object bounding box, the text is assigned to that object. This step effectively binds semantic content to each flowchart element.

\subsection{Arrow Association}

We then associate detected \textit{Arrow} with their corresponding start and end points. An \textit{Arrow} is linked to an \textit{Arrow Start} and \textit{Arrow End} based on two criteria:
\begin{enumerate}
    \item The \textit{Arrow Start} and \textit{Arrow End} must be located near the edges of the \textit{Arrow}'s bounding box.
    \item The \ac{iou} between the bounding box formed by the \textit{Arrow Start} and \textit{Arrow End} and the detected \textit{Arrow}'s bounding box must exceed 0.5.
\end{enumerate}

This matching process enables us to recover the directional information inherent in flowcharts. Additionally, textual annotations such as ``yes'' or ``no'' that are not directly associated with any object but are located near an \textit{Arrow} are attached to that \textit{Arrow}.

\subsection{Linking Objects and Arrows}

Once arrows have been associated with their start and end points, we link non-arrow objects (e.g., processes, decisions) to arrows.

For each non-arrow object, we associate any \textit{Arrow Start} located near its bounding box edges as an outgoing connection, and any \textit{Arrow End} located near its edges as an incoming connection. This step reconstructs the underlying control flow or decision logic of the diagram.

\subsection{Prompt Construction}

Using the extracted text, object categories, and relational information, we generate structured prompts that represent the recovered graph structure. For each object, the prompt encodes:
\begin{enumerate}
    \item The object category (e.g., process, decision)
    \item The object's text content
    \item The preceding steps (connected via incoming arrows)
    \item The subsequent steps (connected via outgoing arrows)
\end{enumerate}

These graph-aware prompts are designed to make explicit the topology that is implicit in the visual layout of the flowchart.

\subsection{Question Generation and \ac{vlm} Inference}

Finally, we formulate two types of input for the GPT-4o \ac{vlm}: one without explicit graph information and one incorporating the constructed graph prompts. For each test flowchart, we generate three types of questions:
\begin{enumerate}
    \item \textbf{Next-step prediction:} \textit{In this flowchart diagram, what is the next step after 'xxx'?}
    \item \textbf{Conditional branch prediction:} \textit{In this flowchart diagram, if 'xxx' is 'yyy', what is the next step?}
    \item \textbf{Preceding-step discrimination:} \textit{In this flowchart diagram, which of the steps before 'xxx' except 'zzz'?}
\end{enumerate}

We pass these questions along with the relevant flowchart prompt to the VLM and retrieve its answers. Answer correctness is determined by comparing the VLM's response against a human-annotated ground-truth answer set, with the verification itself handled via an additional LLM-assisted comparison step.

%% file: sections/results.tex
\section{Results}

\subsection{Effectiveness of \ac{ocr} and Detection Model in FlowchartQA}

We compared two approaches for flowchart-based question answering (FlowchartQA): (1) an \ac{ocr} and detection model combination (Model Ocr-Dec) and (2) a no-\ac{ocr} and no-detection baseline using only raw images (Model No-Ocr-Dec). On a specially annotated corpus of 90 questions, we evaluated how explicitly recovering arrow directions and node connections impacts overall QA accuracy.

\subsection{Experimental Setup}

We conducted experiments on a manually annotated corpus consisting of 30 flowchart diagrams. Each diagram was associated with three types of questions, totaling 90 questions across different diagram sizes (Large, Medium, and Small). The detailed settings are summarized in Table~\ref{tab:setup}.

\begin{table}[h]
\centering
\caption{Summary of Experimental Settings}
\label{tab:setup}
\resizebox{\linewidth}{!}{%
\begin{tabular}{ll}
\hline
Item & Details \\
\hline
Corpus & 30 manually annotated flowcharts (10 Large / 10 Medium / 10 Small). \\
       & Each diagram paired with three types of questions, totaling 90 questions. \\
Question Types & Type 1: Next Step; Type 2: Conditional Branch; Type 3: Previous Step \\
Size Categories & Large ($>$22 arrows), Medium (13--22 arrows), Small ($<$13 arrows) \\
Model Ocr-Dec & \ac{ocr} + Detection: Azure AI Document Intelligence \ac{ocr} + DAMO-YOLO object detector. \\
       & Structured prompt and image input to GPT-4o. \\
Model No-Ocr-Dec & Baseline: Direct prompt and image input to GPT-4o (no \ac{ocr}, no detection). \\
Evaluation Metric & Primarily human evaluation, supplemented by \ac{llm}-based scoring. \\
\hline
\end{tabular}
}
\end{table}

To evaluate the correctness of answers generated by the \ac{llm}, we compared them with manually prepared ground-truth answers using two methods: human judgment (primary) and \ac{llm}-based evaluation (reference).

For the human evaluation, correctness was determined by comparing the predicted object 
\textit{B} in the flowchart with the ground-truth object described as "A is B." The evaluation was case-insensitive and ignored punctuation such as periods.

For the \ac{llm}-based evaluation, we used GPT-4o to assess the semantic similarity between the \ac{llm}\'s response and the reference answer. A prompt was designed to determine whether the two answers were essentially equivalent in meaning.

\subsection{Overall Accuracy}

Table~\ref{tab:overall-accuracy} summarizes the overall accuracy across all 90 questions, aggregating results from Type 1, Type 2, and Type 3. Human evaluation is treated as the primary metric, while automatic scoring using a \ac{llm} is provided for reference.

\begin{table}[htbp]
\small
\centering
\caption{Overall accuracy (\%) and raw counts across all question types (n = 90).}
\label{tab:overall-accuracy}
\begin{tabular}{
  l
  S r@{\,/\,}l
  S r@{\,/\,}l
  S r@{\,/\,}l
  S r@{\,/\,}l
}
\toprule
\multirow{2}{*}{\textbf{Question Type}} &
\multicolumn{3}{c}{\makecell{Ocr--Dec\\(Human)}} &
\multicolumn{3}{c}{\makecell{No-Ocr--Dec\\(Human)}} &
\multicolumn{3}{c}{\makecell{Ocr--Dec\\(LLM)}} &
\multicolumn{3}{c}{\makecell{No-Ocr--Dec\\(LLM)}} \\
\cmidrule(lr){2-4}\cmidrule(lr){5-7}\cmidrule(lr){8-10}\cmidrule(lr){11-13}
& \multicolumn{1}{c}{\%} & \multicolumn{2}{c}{n/N}
& \multicolumn{1}{c}{\%} & \multicolumn{2}{c}{n/N}
& \multicolumn{1}{c}{\%} & \multicolumn{2}{c}{n/N}
& \multicolumn{1}{c}{\%} & \multicolumn{2}{c}{n/N} \\
\midrule
All (Total) & 88.9 & 80 & 90 & 80.0 & 72 & 90 & 78.9 & 71 & 90 & 75.6 & 68 & 90 \\
\bottomrule
\end{tabular}
\end{table}

\subsection{Accuracy by Question Type}

Table~\ref{tab:type-accuracy} summarizes the accuracy results for each question type.  

\begin{table}[htbp]
\small
\centering
\caption{Accuracy (\%) and raw counts for each question type.}
\label{tab:type-accuracy}
\begin{tabular}{
  l
  S r@{\,/\,}l
  S r@{\,/\,}l
  S r@{\,/\,}l
  S r@{\,/\,}l
}
\toprule
\multirow{2}{*}{\textbf{Question Type}} &
\multicolumn{3}{c}{\makecell{Ocr--Dec\\(Human)}} &
\multicolumn{3}{c}{\makecell{No-Ocr--Dec\\(Human)}} &
\multicolumn{3}{c}{\makecell{Ocr--Dec\\(LLM)}} &
\multicolumn{3}{c}{\makecell{No-Ocr--Dec\\(LLM)}} \\
\cmidrule(lr){2-4}\cmidrule(lr){5-7}\cmidrule(lr){8-10}\cmidrule(lr){11-13}
& \multicolumn{1}{c}{\%} & \multicolumn{2}{c}{n/N}
& \multicolumn{1}{c}{\%} & \multicolumn{2}{c}{n/N}
& \multicolumn{1}{c}{\%} & \multicolumn{2}{c}{n/N}
& \multicolumn{1}{c}{\%} & \multicolumn{2}{c}{n/N} \\
\midrule
Type 1 (Next Step)      & 100.0   & 30 & 30 &  83.3 & 25 & 30 &  93.3 & 28 & 30 &  76.7 & 23 & 30 \\
Type 2 (Cond.\ Branch)  &  90.0 & 45 & 50 &  82.0 & 41 & 50 &  84.0 & 42 & 50 &  86.0 & 43 & 50 \\
Type 3 (Previous Step)  &  50.0 &  5 & 10 &  60.0 &  6 & 10 &  10.0 &  1 & 10 &  20.0 &  2 & 10 \\
\bottomrule
\end{tabular}
\end{table}

\subsection{Accuracy by Diagram Size}

Table~\ref{tab:size-accuracy} shows the accuracy categorized by diagram size.  
Again, human evaluation is treated as primary, with \ac{llm} automatic scoring shown for reference.

\begin{table}[htbp]
\small
\centering
\caption{Accuracy  (\%) by diagram size with supporting counts.}
\label{tab:size-accuracy}
\setlength{\tabcolsep}{8pt}
\begin{tabular}{
  @{}l
  S r@{\,/\,}l
  S r@{\,/\,}l
  S r@{\,/\,}l
  S r@{\,/\,}l@{}
}
\toprule
\multirow{2}{*}{\textbf{Diagram Size}} &
\multicolumn{3}{c}{\makecell{Ocr--Dec\\(Human)}} &
\multicolumn{3}{c}{\makecell{No-Ocr--Dec\\(Human)}} &
\multicolumn{3}{c}{\makecell{Ocr--Dec\\(\ac{llm})}} &
\multicolumn{3}{c}{\makecell{No-Ocr--Dec\\(\ac{llm})}} \\
\cmidrule(lr){2-4}\cmidrule(lr){5-7}\cmidrule(lr){8-10}\cmidrule(lr){11-13}
& \multicolumn{1}{c}{\%} & \multicolumn{2}{c}{n/N}
& \multicolumn{1}{c}{\%} & \multicolumn{2}{c}{n/N}
& \multicolumn{1}{c}{\%} & \multicolumn{2}{c}{n/N}
& \multicolumn{1}{c}{\%} & \multicolumn{2}{c}{n/N} \\
\midrule
Large  & 80.0 & 24 & 30 & 66.7 & 20 & 30 & 63.3 & 19 & 30 & 50.0 & 15 & 30 \\
Medium & 93.3 & 28 & 30 & 80.0 & 24 & 30 & 80.0 & 24 & 30 & 80.0 & 24 & 30 \\
Small  & 93.3 & 28 & 30 & 93.3 & 28 & 30 & 93.3 & 28 & 30 & 96.7 & 29 & 30 \\
\bottomrule
\end{tabular}
\end{table}

%% file: sections/discussion.tex
\section{Discussion}

The experimental results revealed several important insights.  
First, for Type 1 (Next Step) questions, the \ac{ocr} and detection model achieved perfect accuracy (100\%) according to human evaluation, significantly outperforming the no-\ac{ocr}-Dec baseline by 16.7 percentage points. \ac{llm}-based scoring similarly showed large gains (+16.7 pp), validating the robustness of this improvement.

For Type 2 (Conditional Branch) questions, Model \ac{ocr}-Dec improved by 8.0 percentage points based on human evaluation, though \ac{llm} automatic scoring showed almost no advantage. This discrepancy suggests that minor variations in textual explanations, which human evaluators can tolerate, may cause automatic scorers to incorrectly penalize correct answers.

For Type 3 (Previous Step) questions, both human and \ac{llm} evaluations revealed low accuracy, with Model no-\ac{ocr}-Dec slightly outperforming Model \ac{ocr}-Dec. This confirms that execution order reasoning remains difficult without explicit graph structure input.

Regarding diagram size, Model \ac{ocr}-Dec outperformed the baseline on Large and Medium diagrams in human evaluations. Improvements were smaller or absent for Small diagrams, which tend to have simpler structures where explicit arrow recovery has less impact.

\subsection{Error Analysis and Improvement Strategies}

Error analysis highlighted several recurring failure patterns.  
A primary source of error was mislinking of arrow endpoints, sometimes connecting decision branches (e.g., ``Yes''/``No'') incorrectly. Introducing an \ac{iou}-based post-correction method after detection is expected to address this issue.

Another common error was \ac{ocr} over-segmentation, where contiguous phrases were split into multiple fragments. Distance-based clustering of bounding boxes could help merge these fragmented texts.

Furthermore, failure to recover complete graph topology, particularly when nodes had multiple incoming edges, often led to incorrect reasoning. Representing the flowchart as a JSON-encoded directed graph, with topological ordering explicitly embedded in prompts, is a promising solution.

Finally, it should be emphasized that \ac{llm} automatic scoring showed limitations in handling paraphrases and extended explanations. Therefore, human evaluation was adopted as the principal measure of accuracy, and \ac{llm} results were treated as supplementary indicators.

%% file: sections/conclusion.tex
\section{Conclusion}

This study demonstrated that combining \ac{ocr} and flowchart-specific object detection substantially improves question answering accuracy for flowcharts, particularly in large diagrams and next-step reasoning tasks (Type 1). By explicitly recovering text content and arrow directions, the proposed method enabled \ac{llm}s to better understand the structural relationships embedded in flowchart diagrams.

Evaluation was primarily conducted via human judgment, supplemented by automatic scoring using a secondary \ac{llm}. Human evaluation revealed that the \ac{ocr} and detection model achieved perfect accuracy for next-step questions (Type 1) and substantial improvements for conditional branch questions (Type 2), confirming the effectiveness of explicitly structured input.  
However, \ac{llm} automatic evaluation sometimes underreported accuracy, especially when model outputs included extended explanations, highlighting the limitations of strict string-matching approaches for complex reasoning tasks.

While significant gains were observed for next-step questions, challenges remain for conditional branching (Type 2) and previous-step identification (Type 3). In these cases, simple text extraction and object localization were insufficient; fine-grained understanding of control flow, decision logic, and execution order is critical. Further improvements will require:

\begin{itemize}
    \item High-precision detection of arrow start and end points to prevent directional ambiguity
    \item Explicit representation of the flowchart\'s graph structure in prompts, allowing the \ac{llm} to reason over paths and dependencies
\end{itemize}

Moreover, the error analysis highlighted additional areas for refinement, such as mitigating \ac{ocr} over-segmentation errors and incorporating graph-based topological information directly into the reasoning pipeline.  
Addressing these challenges is expected not only to boost performance on complex reasoning tasks but also to improve system robustness when applied to handwritten diagrams, \ac{bpmn}, and industrial schematics.

Finally, the modular pipeline proposed here—separating visual parsing from reasoning—paves the way for scalable, domain-adaptive flowchart understanding systems. Future work will explore enhancing graph-structured prompting, developing confidence-aware reasoning mechanisms, and improving automatic evaluation methods to better handle paraphrastic or explanatory outputs, thus enabling more reliable and generalizable deployment across diverse real-world settings.

%% file: sections/appendix.tex
\section{Additional Evaluation Results}

We provide here additional results and analysis that complement the main paper, including per-category detection performance and relaxed \ac{iou} evaluations.

\subsection{Detection Results}
We evaluated the detection performance of the DAMO-YOLO model on our custom test dataset using the COCO evaluation metrics. Table~\ref{tab:overall_ap_ar} shows the \ac{ap} and \ac{ar} across different object sizes under relaxed \ac{iou} thresholds (0.10–0.50). The overall \ac{ap} was 0.836 and \ac{ar} reached 0.925, with large objects achieving the highest recall (\ac{ar} = 0.984).

\begin{table}[!htb]
  \centering
  \caption{Overall \ac{ap} and \ac{ar} (\ac{iou}=0.10–0.50) for different object sizes}
  \label{tab:overall_ap_ar}
  \begin{tabular}{lcccc}
    \toprule
    Metric & All & Small & Medium & Large \\
    \midrule
    \ac{ap}@0.10–0.50 & 0.836 & 0.785 & 0.832 & 0.831 \\
    \ac{ar}@maxDets=100 & 0.925 & 0.897 & 0.872 & 0.984 \\
    \bottomrule
  \end{tabular}
\end{table}

Table~\ref{tab:category_ap} reports category-wise \ac{map} under the standard COCO setting (\ac{iou} = 0.50–0.95). The \textit{Arrow} class achieved moderate performance (\ac{map} = 0.4476). However, the average \ac{map} for all arrow-related categories including \textit{Arrow Start} and \textit{Arrow End} was significantly lower (\ac{map} = 0.2349) compared to non-arrow categories (\ac{map} = 0.6531).

\begin{table}[!htb]
  \centering
  \caption{Per-category \ac{map} (\ac{iou}=0.50–0.95)}
  \label{tab:category_ap}
  \begin{tabular}{lc}
    \toprule
    Category & \ac{map} \\
    \midrule
    \textit{Arrow} & 0.4476 \\
    Arrow-related (\textit{Arrow}, \textit{Arrow Start}, \textit{Arrow End}) & 0.2349 \\
    Non-arrow categories & 0.6531 \\
    All categories & 0.5137 \\
    \bottomrule
  \end{tabular}
\end{table}

Since the bounding boxes for \textit{Arrow Start} and \textit{Arrow End} are very small, their detection accuracy tends to be underestimated when evaluated with the standard \ac{iou} range of 0.50–0.95. Therefore, we also evaluated them under a lower \ac{iou} range of 0.10–0.50. The results are shown in Table~\ref{tab:lowiou_ap}.

\begin{table}[!htb]
  \centering
  \caption{\ac{map} of small objects under relaxed \ac{iou} (0.10–0.50)}
  \label{tab:lowiou_ap}
  \begin{tabular}{lc}
    \toprule
    Category & \ac{map}@0.10–0.50 \\
    \midrule
    \textit{Arrow Start} & 0.7541 \\
    \textit{Arrow End}   & 0.8373 \\
    \bottomrule
  \end{tabular}
\end{table}

\section{LLM-as-a-Judge Evaluation Details}
\label{sec:llm_judge}

For the \ac{llm}-based evaluation described in the main paper, we used the following prompt to assess the similarity between model-generated answers and reference answers:

\begin{lstlisting}[breaklines=true, basicstyle=\small\ttfamily, frame=single, columns=flexible]
You are a strict judge tasked with the following:

1. A question (Question)
2. A reference answer (Reference Answer)
3. A model output (Model Output)

Please evaluate the model output by following these steps:

### Step 1: Analyze the Answers
- First, compare the reference answer and the model output.
- Determine whether they essentially match in meaning or reasoning, or if the model output is otherwise correct based on its logic and evidence.
- Provide a thorough and logical assessment, noting any gaps or inconsistencies.

### Step 2: Final Judgment
- If the model output is substantially the same as the reference answer or equivalently valid judge it as correct.
- If there are clear mistakes, omissions, or inconsistencies, judge it as incorrect.

### Step 3: Output in the Specified Schema
- Please output your evaluation result strictly in the following JSON format:
\end{lstlisting}

Where [Reference Answer] and [LLM Answer] were replaced with the actual reference and LLM-generated answers, respectively. We also utilized Structured Outputs to ensure consistent formatting of the evaluation results in JSON format, making the automated processing of judgments more reliable.